\documentclass[letterpaper, 10pt, conference]{ieeeconf}

\IEEEoverridecommandlockouts
\makeatletter
\def\endfigure{\end@float}
\def\endtable{\end@float}

\makeatother

\usepackage{cite}

\usepackage{times}
\usepackage{amsmath,amssymb,amsfonts}

\usepackage{graphicx}      
\usepackage[table]{xcolor}
\usepackage{array}
\usepackage{booktabs}
\usepackage{multirow}
\usepackage{makecell}

\usepackage{algorithm}
\usepackage{algpseudocode}

\usepackage{etoolbox}
\usepackage{balance}
\usepackage{hyperref}

\usepackage[caption=false,font=footnotesize]{subfig}

\newcommand{\second}[1]{\textcolor{black!50}{\textbf{#1}}}
\newcommand{\OURS}{C$^2$-Explorer}

\makeatletter
\newif\if@introfigdone
\@introfigdonefalse

\let\@oldmaketitle\@maketitle
\renewcommand{\@maketitle}{%
  \@oldmaketitle
  \if@introfigdone\else
    \begingroup
      \centering
      \includegraphics[width=0.99\textwidth]{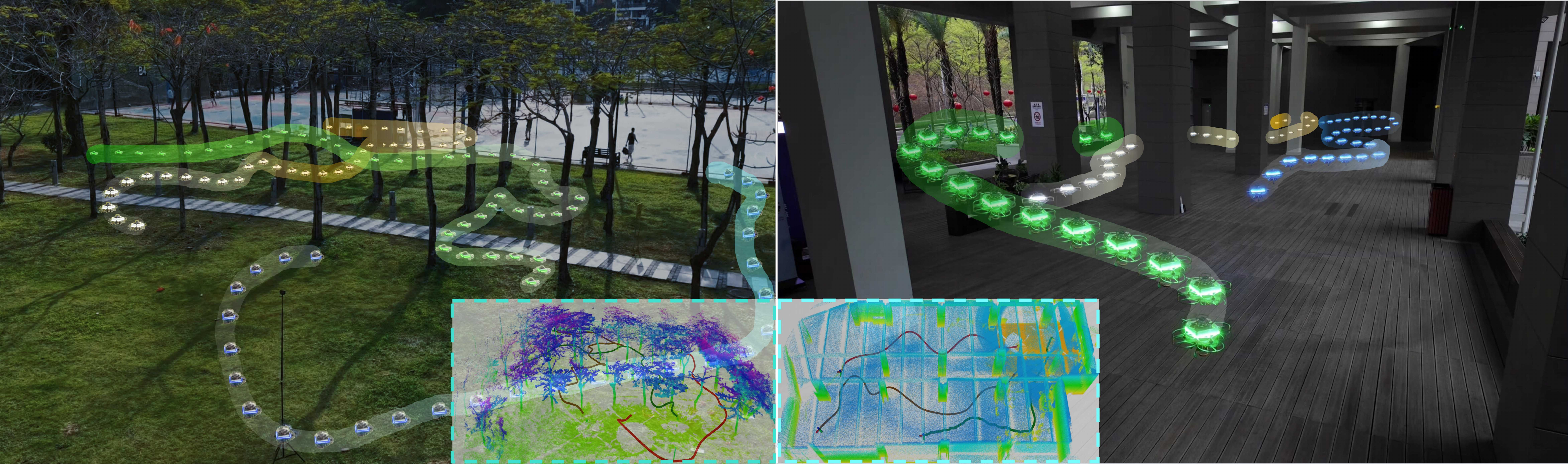}
      \vspace{-0.1cm}
      \def\@captype{figure}%
      \caption{\textbf{Real-World Demonstration of the 3-UAV \OURS}. The system explores an unstructured woodland with trees (left) and a structured building with pillars (right). Inset images display the corresponding generated point cloud maps and flight trajectories.}
      \label{fig:intro}
      \vspace{-0.2cm}
    \endgroup
    \global\@introfigdonetrue
  \fi
}
\makeatother

\title{\fontsize{15.9}{20}\selectfont\bfseries
C$^2$-Explorer:
Contiguity-Driven Task Allocation with Connectivity-Aware Task Representation for Decentralized Multi-UAV Exploration
}

\author{Xinlu Yan$^{1, \textbf{*}}$, Mingjie Zhang$^{2,3, \textbf{*}}$, Yuhao Fang$^{1}$, Yanke Sun$^{1}$, \\
Jun Ma$^{2}$, Youmin Gong$^{1}$, Boyu Zhou$^{3, \dag}$, Jie Mei$^{1, \dag}$ \\ \textbf{* Equal Contribution} \textbf{$^{\dag}$ Co-corresponding Author} 
\thanks{$^{1}$ School of Intelligence Science and Engineering, Harbin Institute of Technology, Shenzhen, Guangdong, China}
\thanks{$^{2}$ Robotics and Autonomous Systems Thrust, The Hong Kong University of Science and Technology (Guangzhou), Guangzhou, Guangdong, China}%
\thanks{$^{3}$ Department of Mechanical and Energy Engineering, Southern University of Science and Technology, Shenzhen, Guangdong, China}%
\thanks{{\tt\footnotesize 24s153069@stu.hit.edu.cn}, {\tt\footnotesize zagerzhang@gmail.com}}
\thanks{{\tt\footnotesize zhouby@sustech.edu.cn}, {\tt\footnotesize jmei@hit.edu.cn}}
}
\begin{document}
\maketitle

\thispagestyle{empty}
\pagestyle{empty}
\begin{abstract}
Efficient multi-UAV exploration under limited communication is severely bottlenecked by inadequate task representation and allocation. Previous task representations either impose heavy communication requirements for coordination or lack the flexibility to handle complex environments, often leading to inefficient traversal. Furthermore, short-horizon allocation strategies neglect spatiotemporal contiguity, causing non-contiguous assignments and frequent cross-region detours. To address this, we propose C$^2$-Explorer, a decentralized framework that constructs a connectivity graph to decompose disconnected unknown components into independent task units. We then introduce a contiguity-driven allocation formulation with a graph-based neighborhood penalty to discourage non-adjacent assignments, promoting more contiguous task sequences over time. Extensive simulation experiments show that C$^2$-Explorer consistently outperforms state-of-the-art (SOTA) baselines, reducing average exploration time by 43.1\% and path length by 33.3\%. Real-world flights further demonstrate the system's feasibility. The code will be released at \href{https://github.com/Robotics-STAR-Lab/C2-Explorer}{\textcolor{blue}{https://github.com/Robotics-STAR-Lab/C2-Explorer}}.
\end{abstract}

\vspace{-0.0cm}
\section{INTRODUCTION}

Autonomous exploration with unmanned aerial vehicles (UAVs) is critical for various applications, such as rapid 3D reconstruction\cite{zhang2024soar}, search-and-rescue\cite{petrlik2025uav, zhang2025apexnav}, and structural inspection\cite{hollinger2013active, cao2025caric}. However, single-UAV exploration is hampered by insufficient flight endurance, resulting in limited coverage capability. To achieve wider coverage, recent works \cite{zhou2023racer, cao2023representation, bartolomei2023fast} have increasingly focused on multi-UAV exploration. Nevertheless, realizing effective team collaboration for efficient exploration remains a significant challenge.

Appropriate task allocation is crucial for efficient multi-UAV exploration, as it ensures workload balance and avoids redundant task assignments. However, existing methods remain limited by bottlenecks in task representation and spatiotemporal contiguity in task allocation. 
On the one hand, many approaches \cite{bartolomei2023fast, zhang2024leces, hui2025pc, lewis2025frontier} employ frontiers or viewpoints as task units, which demand real-time globally consistent map sharing among UAVs and impose strict requirements on communication. In large-scale environments with frequent occlusions, such seamless map sharing is often impractical. While some works \cite{zhou2023racer, cao2023representation} adopt region-based task representation via space partitioning to mitigate this issue, they rely on uniform grid decomposition that assumes regular structure, leading to degraded performance in structurally complex spaces.
On the other hand, current methods fail to satisfy the critical requirement of spatiotemporal contiguity, which refers to the consistent assignment of spatially adjacent tasks to a single UAV over consecutive time steps, thereby reducing inefficient cross-regional traversal. Specifically, greedy methods \cite{hui2025pc, lewis2025frontier, dong2024fast} select a locally optimal viewpoint for each UAV at each decision instant, ignoring temporal consistency. This leads to frequent switching among spatially disjoint targets and degrades operational efficiency. Although methods \cite{zhou2023racer, zhang2024leces, gao2022meeting} account for inter-regional traversal costs, they lack explicit contiguity regularization. Consequently, they may still assign disconnected tasks to individual UAVs, making contiguous task subsets difficult to maintain over time.

\begin{figure*}[t]
  \vspace{0.0cm} 
  \centering
  \includegraphics[width=0.99\textwidth]{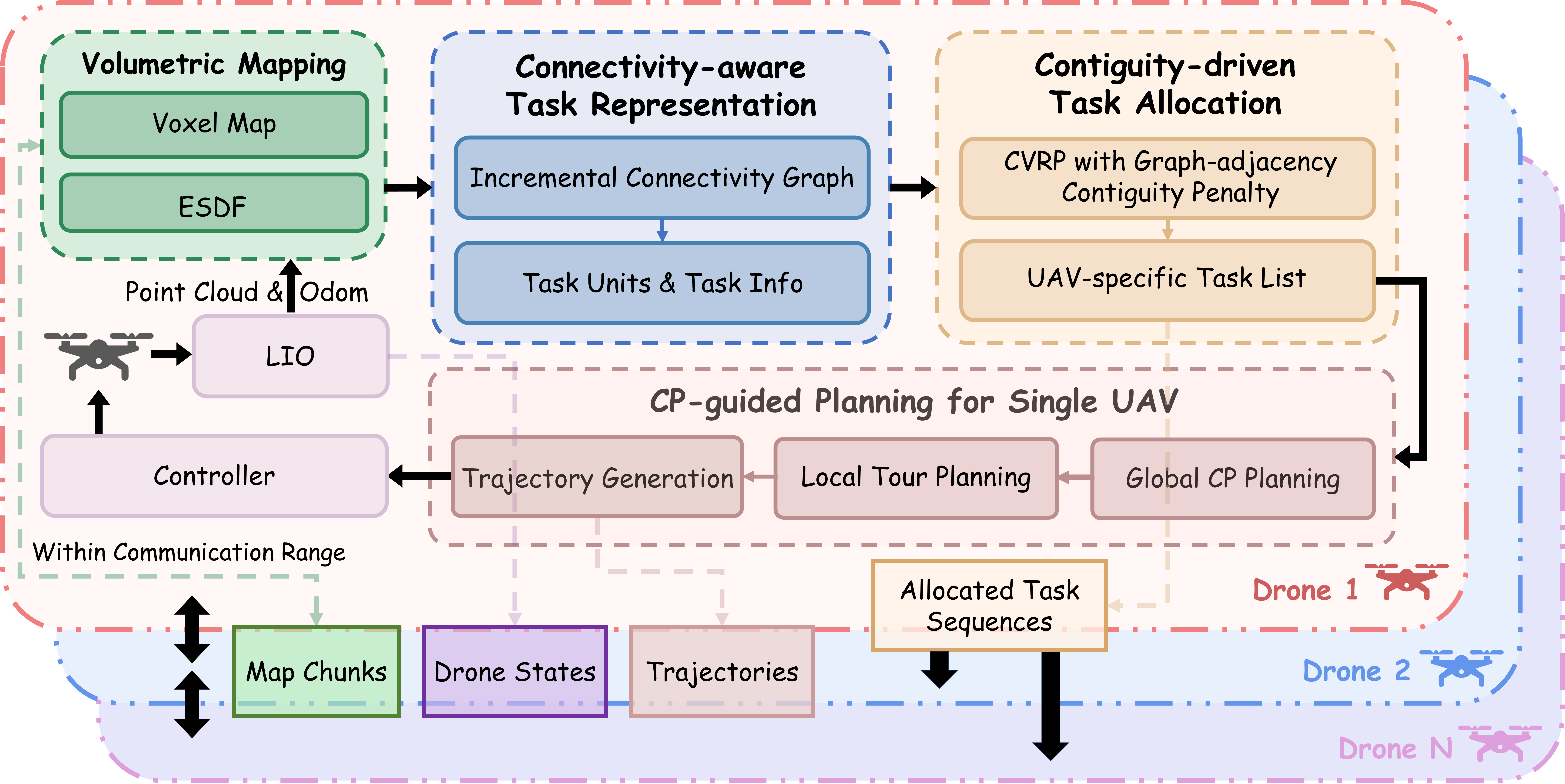}
  \vspace{-0.10cm}
  \caption{\textbf{System Architecture of \OURS.} Within communication range, Drone 1 (the lowest-ID UAV) acts as a temporary host. It builds volumetric maps via LIO and extracts task units using a connectivity-aware representation. The host then solves a contiguity-regularized CVRP to generate per-UAV task sequences and disseminates them to the other drones. Each drone subsequently executes CP-guided planning for navigation. In parallel, drones exchange local map updates, states, and trajectories within communication range to reduce redundant exploration and support collision avoidance. }
  
  \label{fig:overview}
  \vspace{-0.30cm}
\end{figure*}

To address the limitations, we present \textbf{\OURS}, a decentralized multi-UAV exploration framework designed to enhance flexibility and contiguity in task allocation. First, we introduce a connectivity-aware task representation to address the
topology-agnostic limitation of grid-based partitioning in structurally complex environments. Specifically, we build a connectivity graph upon a coarse grid partitioning and split spatially disconnected unknown regions within each grid into independent task units. This avoids unnecessary detours caused by merging disjoint regions into one task, yielding more reasonable task units for efficient task allocation. To overcome the lack of spatiotemporal contiguity in traditional allocation strategies, we formulate an optimization problem that accounts for both the total traversal distance of the UAV team and allocation adjacency constraints. Specifically, we leverage topological adjacency information from the connectivity graph to penalize task assignments that span spatially disconnected regions. This mechanism promotes grouping of spatially connected tasks and improves allocation consistency across replanning steps, reducing interleaved trajectories and cross-region detours.

We validate \OURS\ through extensive simulations and real-world experiments. In simulation benchmarks, \OURS\ reduces exploration time by \textbf{43.1\%} and total path length by \textbf{33.3\%} on average compared to SOTA baselines. We further deploy \OURS\ on real UAV platforms to demonstrate its practical effectiveness and robustness. In summary, our key contributions are

1) A connectivity-aware task representation that models connectivity over the partitioned space and splits spatially disconnected unknown regions into independent task units, reducing unnecessary traversal and improving efficiency.

2) A contiguity-driven task allocation, formulated as an optimization over traversal cost with adjacency constraints, penalizing disconnected assignments to encourage spatially and temporally contiguous task sets.

3) Extensive validation in simulation and real-world experiments demonstrating the effectiveness and efficiency of \OURS. The source code will be released.

\section{RELATED WORK}
\subsection{Single-UAV Exploration}
Autonomous exploration has been widely studied, and classical methods can be roughly categorized into sampling-based \cite{bircher2016receding, yang2021graph, duberg2022ufoexplorer} and frontier-based methods \cite{yamauchi1997frontier, cieslewski2017rapid}. However, these methods are often locally greedy, causing back-and-forth motion and redundant revisits. Hierarchical frameworks \cite{zhou2021fuel, yu2023echo, bu2025rush} introduce global guidance to enhance consistency but may overlook the underlying environment topology, leading to occasional deviations from the intended global guidance. Conversely, recent work \cite{zhang2025falcon} incrementally builds a connectivity graph to explicitly encode topological structure for more reliable coverage path (CP) guidance. Nevertheless, single-UAV exploration is inherently constrained by insufficient flight endurance, motivating increasing interest in multi-UAV collaborative exploration.

\subsection{Task Representation for Multi-UAV Exploration}

In multi-UAV exploration, an effective task representation is fundamental to flexible task allocation. Many approaches \cite{bartolomei2023fast, zhang2024leces, hui2025pc, lewis2025frontier, gao2022meeting} adopt frontiers or viewpoints as task units, abstracting exploration objectives as frontier-derived targets or informative viewpoints. While \cite{bartolomei2023fast, zhang2024leces, lewis2025frontier} group adjacent frontiers into clusters, Hui \textit{et al}.\cite{hui2025pc} condense task-related information into the viewpoints sampled around frontier clusters. To reduce the scale of task allocation, Gao \textit{et al}.\cite{gao2022meeting} integrate spatially proximal viewpoints into super viewpoints. However, these representations rely on continuous global map sharing, which becomes a critical limitation in large-scale environments with limited communication and frequent occlusions. 
To address this, some works \cite{zhou2023racer, cao2023representation, dong2024fast} embrace region-based task representation by partitioning the exploration into subregions. Zhou \textit{et al}.\cite{zhou2023racer} decompose the unknown space into hierarchical grids, utilizing disjoint cells of varying resolutions as elementary task units. Additionally, Cao \textit{et al}.\cite{cao2023representation} incorporate a dual-resolution representation, with uniform cuboid subspaces as coarse-grained global task units. Dong \textit{et al}.\cite{dong2024fast} introduce Exploration Regions of Interest to represent the uniform decomposition, structured as nodes within a multi-robot dynamic topological graph to facilitate coordination. Nevertheless, such uniform decomposition imposes a rigid geometric structure that is topologically inconsistent with complex environments, resulting in inappropriate task allocation. To tackle this challenge, we propose connectivity-aware task representation that models the connectivity of the decomposed spatial regions and partitions spatially disconnected unknown areas into independent task units, enabling more flexible task allocation.

\subsection{Task Allocation for Multi-UAV Exploration}
Building upon the established task representation, reasonable task allocation is essential to achieving efficient multi-UAV collaborative exploration. 
In early work \cite{yamauchi1999decentralized}, each UAV selects the frontier closest to itself as the next target. To mitigate the allocation of redundant task units, recent approaches have incorporated a repulsion mechanism. Bartolomei \textit{et al}. \cite{bartolomei2023fast} utilize repulsive potential fields to spatially separate UAVs' areas of interest. Moreover, Hui \textit{et al}. \cite{hui2025pc} penalize the utility of overlapping viewpoints to implicitly guide UAVs toward distinct exploration regions. However, relying solely on instantaneous decisions often neglects temporal consistency, leading to frequent target switching. 
Unlike greedy methods, some works \cite{zhou2023racer, cao2023representation, gao2022meeting} formulate task allocation as a Vehicle Routing Problem (VRP). Specifically, Cao \textit{et al}.\cite{cao2023representation} solve a min-max VRP to assign a sequence of sub-regions to each UAV, minimizing the maximum traversal cost. Similarly, Gao \textit{et al}.\cite{gao2022meeting} establish a cost matrix from traversal distances for VRP modeling. In addition to path cost, Zhou \textit{et al}.\cite{zhou2023racer} construct a CVRP with workload as the capacity constraint, and the consistency term is implemented as a constant bias added in the cost matrix. These methods neglect explicit spatial contiguity regularization and thus fail to generate spatially contiguous and temporally consistent task assignments for UAVs. We develop a contiguity-driven task allocation method, formulated as an adjacency-constrained traversal cost optimization that penalizes disconnected assignments to promote spatially contiguous tasks over time.

\begin{figure*}[t]
  \vspace{0.0cm}
  \centering
  \includegraphics[width=0.95\textwidth]{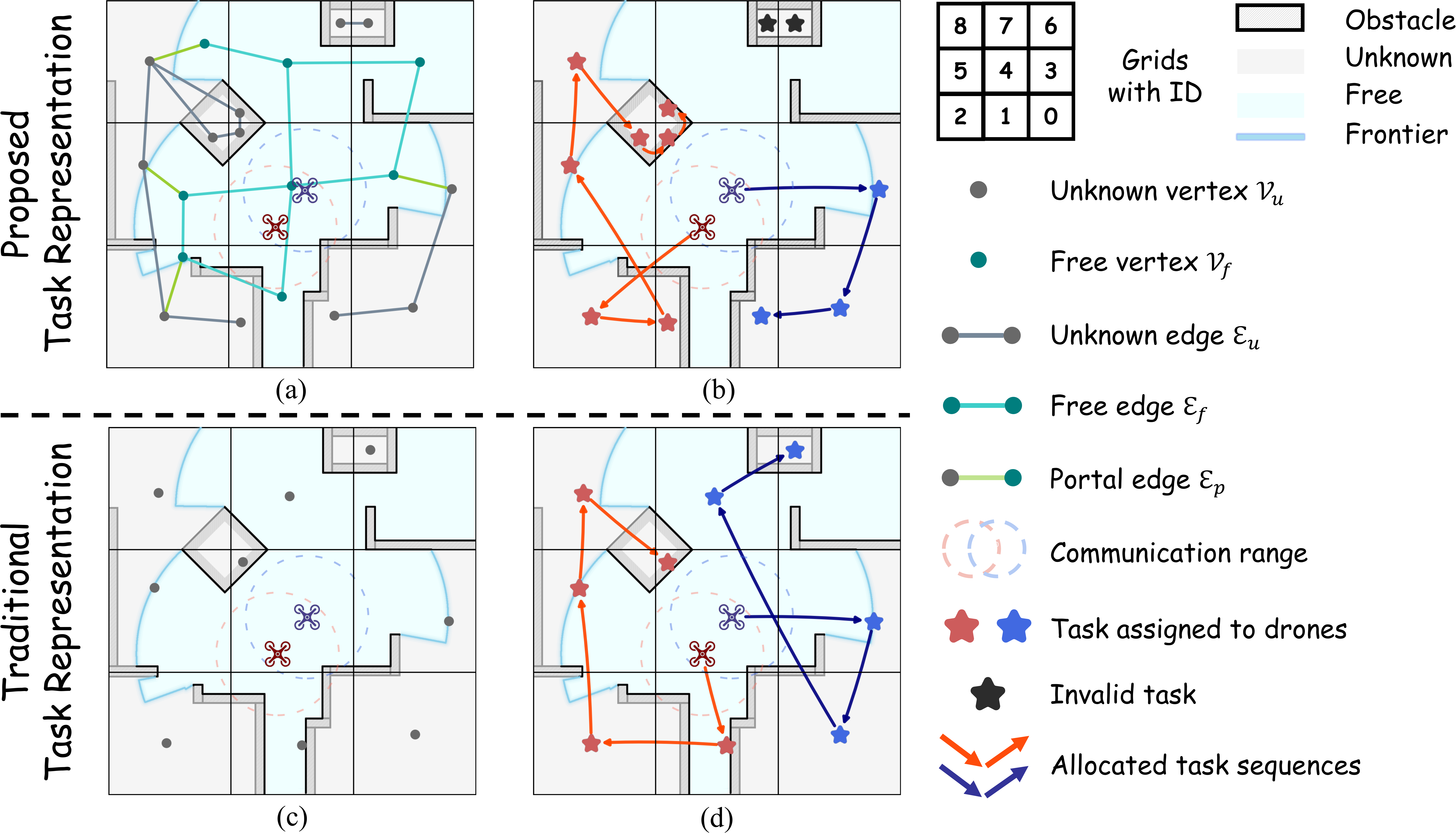}
  \vspace{-0.10cm}
  \caption{\textbf{Comparison of Connectivity-Aware and Traditional Task Representations.} (a) and (c) contrast environmental modeling: (a) shows our connectivity graph modeling the topology of free and unknown spaces, while (c) uses uniform grid decomposition based on unknown voxel centroids. (b) and (d) compare task representation and allocation outcomes: (b) illustrates our connectivity-aware approach, treating spatially disjoint unknown regions as task units and excluding obstacle-enclosed areas. Conversely, (d)'s topology-agnostic approach merges disconnected regions and retains unreachable tasks.}
  \label{fig:methods}
  \vspace{-0.30cm}
\end{figure*}

\section{Problem Formulation}

We consider a decentralized multi-UAV system to explore an unknown and spatially limited 3D space \(V \subset \mathbb{R}^3\) enclosed by a bounding box \(B\). Our system comprises \(N\) UAVs equipped with a LiDAR sensor for rapid exploration. 
UAVs can only communicate with each other within a given distance \(r_{comm}\). When communication is available, UAVs collaboratively allocate tasks and share map chunks\cite{zhou2023racer}. Each UAV becomes idle when no frontier remains in its assigned task units. Exploration is complete only when all valid task units have been completed and no frontier remains globally.

\begin{table}[t]
\vspace{0.0cm}
\caption{Data contained by a $\mathcal{T}_i$ task unit.}
\vspace{-0.10cm}
\label{tab:task unit information structure}
\centering
\renewcommand{\arraystretch}{1.15}
\setlength{\tabcolsep}{10pt}
\begin{tabular*}{0.89\columnwidth}{@{\extracolsep{\fill}} >{\centering\arraybackslash}m{0.25\columnwidth} >{\centering\arraybackslash}m{0.55\columnwidth} @{}}
\hline\hline
\textbf{Data} & \textbf{Explanation} \\
\hline
$\mathbf{p}_{u,i}$ & Unknown vertex position \\
$\mathcal{H}_i$ & 2D convex hull (if split) \\
$g_{i}$ & Parent grid ID (if not split) \\
$\text{NUM}_{i}$ & Number of unknown voxels \\
$\text{d}_i$ & UAV ID \\
\hline\hline
\end{tabular*}
\vspace{-0.30cm}
\end{table}

\section{METHODOLOGY}

As illustrated in Fig.~\ref{fig:overview}, the proposed \OURS\ system consists of three main components. First, each UAV incrementally builds a 3D occupancy map and maintains a topology-aware connectivity graph. This graph organizes the unknown space into flexible task units with online status and workload updates (Sec.~\ref{sec:task_representation}). Second, when within communication range, the UAVs coordinate in a decentralized manner to compute a contiguous task assignment, effectively minimizing redundant exploration and cross-region detours (Sec.~\ref{sec:task_allocation}). Finally, based on the allocated tasks, each UAV executes CP-guided global task sequencing and generates smooth, collision-free trajectories to explore the designated regions (Sec.~\ref{sec:cp_planning}).

\subsection{Connectivity-aware Task Representation}
\label{sec:task_representation}

Traditional region-based task representation \cite{zhou2023racer, cao2023representation} adopts the centroids of unknown voxels within the grid as task units, as shown in Fig.~\ref{fig:methods}(c). However, this topology-agnostic representation overlooks the underlying connectivity of the regions, often leading to unreasonable task units. Specifically, when a grid contains multiple spatially disconnected unknown regions, the resulting centroid merges these disjoint regions into one task (e.g., Grid 1, 5 and 7 in Fig.~\ref{fig:methods}(d)), causing unnecessary cross-region detours during exploration. Moreover, the lack of filtering for obstacle-enclosed unknown voxels causes inherently invalid task units. Grid 6 shown in Fig.~\ref{fig:methods}(d) has already been completely explored, yet the unreachable unknown region within it is still treated as a valid task unit, which compromises the rationality of task allocation and reduces the efficiency of exploration. To enable task units to encode the topological information of the environment and eliminate the ambiguity in task exploration status, we incrementally construct a connectivity graph, based on which we propose a connectivity-aware task representation.

\subsubsection{\textbf{Connectivity Graph Construction}}

Inspired by \cite{zhang2025falcon}, the entire exploration space is first \emph{coarsely} uniformly decomposed into grids. 
Within each grid, the Connected Component Labeling (CCL) is adopted to segment it into disjoint free-space and unknown regions, and the anchor of each region is computed as the average position of its voxels.

Based on these regions, we incrementally construct a connectivity graph $\mathcal{G}=(\mathcal{V},\mathcal{E})=(\mathcal{V}_f \cup \mathcal{V}_u ,\mathcal{E}_f \cup \mathcal{E}_u \cup \mathcal{E}_p)$, as illustrated in Fig.~\ref{fig:methods}(a). The vertex set $\mathcal{V}$ consists of region anchors, divided into free vertices \(\mathcal{V}_f\) and unknown vertices \(\mathcal{V}_u\). All edges \(\mathcal{E}\) are generated via restricted A* searches and fall into three categories: free edges $\mathcal{E}_f$, unknown edges $\mathcal{E}_u$ and portal edges $\mathcal{E}_p$.
Specifically, $\mathcal{E}_f$ and $\mathcal{E}_u$ are updated through restricted A* searches constrained in their corresponding regions across neighboring grids, for $\mathcal{V}_f$ and $\mathcal{V}_u$ respectively. In terms of $\mathcal{E}_p$, A* searches are strictly constrained within the same grid between $\mathcal{V}_f$ and $\mathcal{V}_u$. 

\subsubsection{\textbf{Task Unit Representation}}

Building upon the topological structure encoded in our incrementally constructed connectivity graph, we define each spatially disjoint unknown region as an independent task unit, with its corresponding unknown vertex $\mathcal{V}_u$ serving as the topological representative, as shown in Fig.~\ref{fig:methods}(b). This connectivity-aware task representation endows each task unit with explicit topological connectivity and unambiguous spatial boundaries, based on which we formalize a standardized discrete state specification and a dedicated information structure for each unit. 
Specifically, each task unit maintains a discrete task status $s \in\{\textsc{pending},\,\textsc{completed},\,\textsc{invalid}\}$, with all units initialized to \textsc{pending}. A unit is marked \textsc{completed} when its corresponding unknown region is fully explored, with no remaining unknown vertices or associated frontiers. Unreachable unknown regions connected solely to other unknown vertices are identified via the CCL algorithm, marked \textsc{invalid} as illustrated in Fig.~\ref{fig:methods}(b), and excluded from subsequent task allocation.

To support efficient transmission of task allocation results and lightweight inter-UAV assignment information exchange under limited communication, a dedicated task unit information structure $\mathcal{T}_i$ is instantiated for each task unit during spatial decomposition. It stores the corresponding unknown vertex position $\mathbf{p}_{u,i}$, which supports the evaluation of traversal distance for task allocation and exploration planning. If a grid is split into multiple independent task units, the convex hull $\mathcal{H}_i$ of the associated unknown region is maintained. Rather than transmitting the complete unknown voxels within the region, this convex hull serves as the minimal enclosing shape to approximate the spatial scope of the corresponding task unit, while enabling lightweight inter-UAV communication through its compact geometric encoding. Otherwise, the unique ID $g_{i}$ of the parent grid is stored directly. Furthermore, $\mathcal{T}_i$ records the number $\text{NUM}_{i}$ of unknown voxels in the region to quantify the exploration workload of the task unit and support workload balancing in the subsequent allocation process. In addition, the drone ID $\text{d}_i$ indicates the ownership of the task unit. Data stored by a $\mathcal{T}_i$ is listed in Table \ref{tab:task unit information structure}.

\begin{figure*}[!t]
  \vspace{0.0cm}
  \centering
  \includegraphics[width=0.95\textwidth]{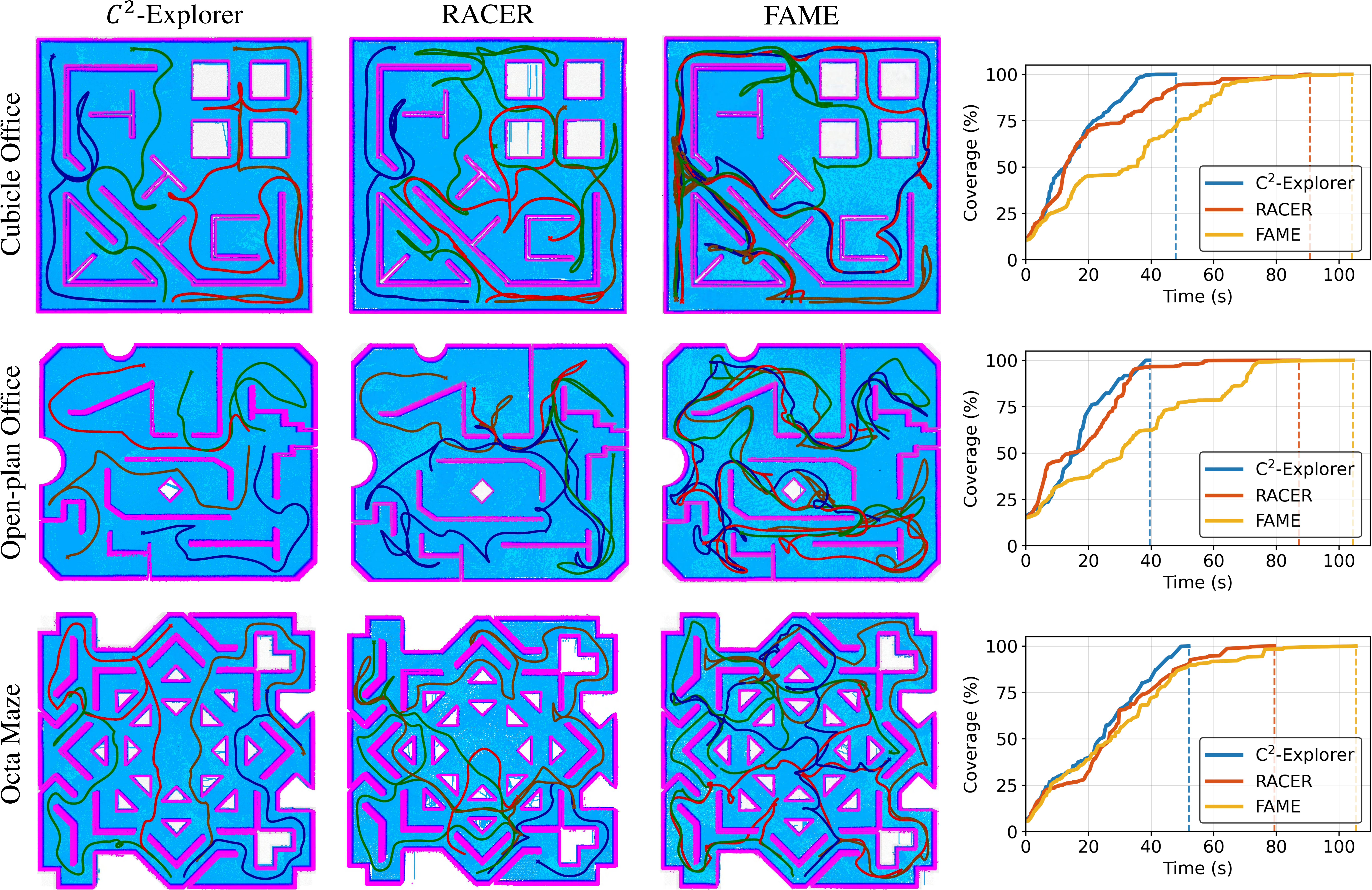}
  \vspace{-0.10cm}
  \caption{\textbf{Benchmark Results of 4-UAV Decentralized Multi-UAV Exploration.} Visualizations of the final executed flight trajectories for \OURS, RACER, and FAME are presented across three complex scenarios: (a) Cubicle Office, (b) Open-plan Office, and (c) Octa Maze. Additionally, the rightmost column presents curves illustrating the evolution of the environment coverage ratio against exploration time for each method.}
  \label{fig:benchmark}
  \vspace{-0.30cm}
\end{figure*}

\subsection{Contiguity-driven Task Allocation}
\label{sec:task_allocation}

\subsubsection{\textbf{Contiguity-driven CVRP}}
Existing multi-UAV exploration task allocation methods \cite{zhou2023racer, gao2022meeting, cao2023representation} only take traversal costs into account in their task assignment optimization. However, these methods lack explicit contiguity regularization in their problem formulation, which tends to assign topologically non-adjacent task units to a single UAV. This will result in interleaved tasks among UAVs, incurring frequent cross-region traversal and overlapping exploration regions, which ultimately degrades the overall exploration efficiency. To address this, we propose contiguity-driven task allocation, leveraging the topological adjacency information from the constructed connectivity graph to penalize non-contiguous task assignments and promote spatially contiguous task grouping for each UAV. 

Following \cite{zhou2023racer}, we formulate the multi-UAV collaborative exploration task allocation problem as a CVRP\cite{zhou2023racer}. Suppose there are $N_c$ UAVs within the communication range, collectively sharing a total of $N_{\mathcal{T}}$ valid task units to be allocated. Therefore, the formulated CVRP involves $N_c + N_{\mathcal{T}} + 1$ nodes, where one additional node is set to the virtual depot. To balance the workload among UAVs, we model the capacity via task-specific demands by setting \(D_i = \text{NUM}_i\) for each task unit. Let \(W \triangleq \sum_{i\in T}\text{NUM}_i\) denote the total workload of all valid task units. The capacity constraint for each UAV is then given by
\begin{equation}
\label{eq:capacity}
    \sum_{i \in T} D_ix_{k,i} \le \frac{\sigma_Q W}{N_c},\quad \forall k\in\{1,\dots,N_c\},
\end{equation}
where \(T\) denotes the set of all valid task units to be allocated, \(x_{k,i}\in\{0,1\}\) is a binary variable indicating whether task unit \(i\) is assigned to \(\text{d}_k\), and \(\sigma_Q \ge 1\) is a capacity scaling factor.
To solve CVRP, the corresponding cost matrix \(\mathbf{C}_{cvrp} \in \mathbb{R}^{(N_c + N_{\mathcal{T}} + 1) \times (N_c + N_{\mathcal{T}} + 1)}\) has the following form:
\begin{equation}
\mathbf{C}_{\mathrm{cvrp}}=
\begin{bmatrix}
M_{inf} & 
\mathbf{0}_{1\times N_c} & 
M_{inf}\mathbf{1}_{1\times N_{\mathcal{T}}} \\
\mathbf{0}_{N_c \times 1} &
\mathbf{0}_{N_c \times N_c} & 
\mathbf{C}_{d,\mathcal{T}} \\
\mathbf{0}_{N_{\mathcal{T}}\times 1} & 
M_{inf}\mathbf{1}_{N_{\mathcal{T}} \times N_c} & 
\mathbf{C}_{\mathcal{T}}
\end{bmatrix}
\label{eq:cvrp}
\end{equation}
where \(M_{inf}\) denotes a sufficiently large positive constant to block infeasible connections. \(\mathbf{C}_{d,\mathcal{T}}\) is a \(N_c \times N_{\mathcal{T}}\) block for UAV-to-task connection cost, while \(\mathbf{C}_{\mathcal{T}}\) represents inter-task connection costs, which is a \({N_{\mathcal{T}} \times N_{\mathcal{T}}}\) block. The two core cost blocks are defined as:
\begin{equation}
\begin{aligned}
& \mathbf{C}_{\mathcal{T}}(i,j) = \mathbf{C}_{\mathcal{T}}(j,i) = \psi(\rho_{ij})l_{i,j}, \\
& \mathbf{C}_{d,\mathcal{T}}(k,i) = \psi(\rho_{d_k,i})l_{d_k,i},
\end{aligned}
\end{equation}
where $i,j \in \{1,\cdots,N_{\mathcal{T}}\}$, $k \in \{1,\cdots, N_c\}$. \(l_{i,j}\) denotes the traversal cost between any two nodes \(\textbf{p}_i\) and \(\textbf{p}_j\), \(\psi(\cdot)\) is the graph-adjacency contiguity penalty kernel, and \(\rho_{ij}\) is the normalized adjacency ratio between node pairs. 

In terms of traversal cost \(l_{i,j}\), we define it as the length of a navigation path computed by a hybrid \(A^*\) strategy:
\begin{equation}
\label{eq:hybrid_astar_cost}
\begin{aligned}
l_{i,j} &\triangleq
\begin{cases}
\operatorname{Len}\!\Big(\mathrm{A}^{*}_{\mathrm{vox}}(\mathbf{p}_i,\mathbf{p}_j;\mathcal{M})\Big),
& \text{if } d_{i,j}<d_{\mathrm{thr}},\\[2pt]
\operatorname{Len}\!\Big(\mathrm{A}^{*}_{\mathrm{graph}}(\Pi(\mathbf{p}_i),\Pi(\mathbf{p}_j);\mathcal{G})\Big),
& \text{if } d_{i,j}\ge d_{\mathrm{thr}}.
\end{cases}
\end{aligned}
\end{equation}
Here,  
$d_{i,j}\triangleq \lVert \mathbf{p}_i-\mathbf{p}_j\rVert_2$ denotes the Euclidean distance.
While \(\mathrm{A}^*_{\text{vox}}(\cdot)\) performs voxel-based \(A^*\) search on the map \(\mathcal{M}\), treating unknown voxels as traversable, \(\mathrm{A}^*_{\text{graph}}(\cdot)\) performs graph-based \(A^*\) search on the connectivity graph \(\mathcal{G}\). The projection operator \(\Pi(\cdot)\) maps a position to its nearest vertex in \(\mathcal{G}\), defined as 
\(\Pi(\mathbf{p}) \triangleq \arg\min_{\mathbf{v}\in\mathcal{V}}\|\mathbf{p} - \mathbf{v}\|_2\).

However, optimizing the multi-UAV task assignment solely based on traversal cost fails to ensure the spatial contiguity of the tasks assigned to each UAV, often yielding unreasonable cross-region assignments and redundant detours that degrade exploration efficiency. Thus, we introduce a graph-adjacency contiguity penalty to penalize non-contiguous task assignments, encouraging connectivity-compliant task grouping. The penalty kernel is given by
\begin{equation}
\begin{aligned}
\label{eq:contiguity penalty}
&\rho_{ij} \triangleq \frac{l_{i,j}}{R_c} = \frac{l_{i,j}}{\lambda_c \cdot L_g}, \\
\psi(\rho_{ij})&=
\begin{cases}
1, & \rho_{ij} \le 1,\\
1+(\rho_{ij}-1)^2, & \rho_{ij} > 1,
\end{cases}
\end{aligned}
\end{equation}
where \(L_g\) denotes the scale of the uniform grid, \(\lambda_c \geq 1\) is a tunable coefficient for the admissible local connectivity range, and \(R_c = \lambda_c \cdot L_g\) is the connectivity radius threshold. The normalized adjacency ratio \(\rho_{ij}\) quantifies the relative separation of a task pair from the allowable neighborhood: \(\rho_{ij} \leq 1\) means the pair lies within the admissible range, while \(\rho_{ij} > 1\) measures its excess beyond this range. 
This kernel applies no additional penalty to topologically adjacent task pairs in the connectivity graph, while introducing a quadratically increasing penalty to pairs outside the admissible neighborhood. Integrating this penalty into the CVRP cost matrix explicitly regularizes assignment continuity, effectively mitigating interleaved trajectories and minimizing redundant cross-region detours.

\subsubsection{\textbf{Commit-Style Task Dispatch}}
Given the CVRP cost matrix $\mathbf{C}_{\mathrm{cvrp}}$, within each group of UAVs that are currently able to communicate, the lowest-ID UAV is selected as a temporary host. The host solves the CVRP and unicasts UAV-specific task sequences $\{\mathcal{S}_k\}_{k=1}^{N_c}$ to all participating UAVs, where $\mathcal{S}_k=(\mathcal{T}_{k,1},\ldots,\mathcal{T}_{k,n_k})$ and $n_k\in\mathbb{Z}_{\ge 0}$, and $N_c$ denotes the number of UAVs in the component.
To maintain consistent task states under lossy and out-of-order links, we adopt a lightweight, timestamped commit-style handshake with three phases:
\emph{Proposal}~$\rightarrow$~\emph{Commit}~$\rightarrow$~\emph{Finalize},
and a \emph{Cancel} message to abort the process.
In \emph{Proposal}, the host broadcasts a versioned allocation (with timestamp) and each UAV validates it, caches its previous assignment, and replies \textsc{Accept}/\textsc{Reject}.
In \emph{Commit}, the host proceeds only after collecting all \textsc{Accept} replies; UAVs then apply the update provisionally and acknowledge.
In \emph{Finalize}, the host confirms activation of the new task lists.
Any rejection or timeout triggers \emph{Cancel}, upon which all UAVs roll back to the cached assignment.
Versioned messages enable idempotent processing of duplicates and late arrivals, while bounded retries with backoff mitigate oscillations caused by frequent reallocations. Overall, this protocol is designed to tolerate packet loss and out-of-order delivery and to keep task states consistent among connected UAVs.

\begin{table}[!t]
\vspace{0.0cm}
\centering
\caption{Comparison of Exploration Performance}
\vspace{-0.10cm}
\label{tab:benchmark}

\resizebox{\columnwidth}{!}{%
\setlength{\tabcolsep}{3pt}
\begin{tabular}{c c c c c c}
\toprule
\textbf{Scene} & \thead{\textbf{Drone}\\\textbf{Num}} & \textbf{Method} &
\thead{\textbf{Exploration}\\\textbf{Time (s)}} &
\thead{\textbf{Total Path}\\\textbf{Length (m)}} &
\thead{\textbf{Flight}\\\textbf{Vel. (m/s)}} \\
\midrule

& & RACER & \second{128.1 $\pm$ 24.1} & \second{372.6 $\pm$ 57.9} & \second{1.84 $\pm$ 0.03} \\
& & FAME  & 142.6 $\pm$ 11.0 & 473.8 $\pm$ 35.9 & 1.71 $\pm$ 0.05 \\
\rowcolor{gray!15} \cellcolor{white} & \cellcolor{white} & & \textbf{73.1 $\pm$ 4.9} & \textbf{253.1 $\pm$ 14.4} & \\
\rowcolor{gray!15} \cellcolor{white} & \cellcolor{white}\multirow{-4}{*}{2} & \multirow{-2}{*}{\textbf{Ours}} & {\scriptsize \textbf{(↓42.9\%)}} & {\scriptsize \textbf{(↓32.1\%)}} & \multirow{-2}{*}{\textbf{1.85 $\pm$ 0.01}} \\
\cmidrule{2-6}

& & RACER & \second{106.2 $\pm$ 8.5} & \second{408.0 $\pm$ 43.3} & \second{1.84 $\pm$ 0.03} \\
& & FAME  & 115.9 $\pm$ 14.8 & 560.5 $\pm$ 62.9 & 1.70 $\pm$ 0.05 \\
\rowcolor{gray!15} \cellcolor{white} & \cellcolor{white} & & \textbf{58.8 $\pm$ 5.6} & \textbf{280.2 $\pm$ 11.4} & \\
\rowcolor{gray!15} \cellcolor{white} & \cellcolor{white}\multirow{-4}{*}{3} & \multirow{-2}{*}{\textbf{Ours}} & {\scriptsize \textbf{(↓44.6\%)}} & {\scriptsize \textbf{(↓31.3\%)}} & \multirow{-2}{*}{\textbf{1.84 $\pm$ 0.02}} \\
\cmidrule{2-6}

& & RACER & \second{90.7 $\pm$ 11.1} & \second{445.8 $\pm$ 74.7} & \second{1.80 $\pm$ 0.06} \\
& & FAME  & 100.8 $\pm$ 20.0 & 614.1 $\pm$ 59.7 & 1.74 $\pm$ 0.07 \\
\rowcolor{gray!15} \cellcolor{white} & \cellcolor{white} & & \textbf{50.9 $\pm$ 4.2} & \textbf{301.7 $\pm$ 14.9} & \\
\rowcolor{gray!15} \cellcolor{white}\multirow{-12}{*}{\shortstack{Cubicle\\Office}} & \cellcolor{white}\multirow{-4}{*}{4} & \multirow{-2}{*}{\textbf{Ours}} & {\scriptsize \textbf{(↓43.9\%)}} & {\scriptsize \textbf{(↓32.3\%)}} & \multirow{-2}{*}{\textbf{1.82 $\pm$ 0.04}} \\
\midrule

& & RACER & \second{125.7 $\pm$ 28.9} & \second{319.3 $\pm$ 54.0} & \second{1.78 $\pm$ 0.07} \\
& & FAME & 169.9 $\pm$ 19.1 & 553.9 $\pm$ 85.5 & 1.67 $\pm$ 0.08 \\
\rowcolor{gray!15} \cellcolor{white} & \cellcolor{white} & & \textbf{67.3 $\pm$ 6.7} & \textbf{214.5 $\pm$ 24.1} & \\
\rowcolor{gray!15} \cellcolor{white} & \cellcolor{white}\multirow{-4}{*}{2} & \multirow{-2}{*}{\textbf{Ours}} & {\scriptsize \textbf{(↓46.5\%)}} & {\scriptsize \textbf{(↓32.8\%)}} & \multirow{-2}{*}{\textbf{1.81 $\pm$ 0.04}} \\
\cmidrule{2-6}

& & RACER & \second{116.1 $\pm$ 19.6} & \second{342.8 $\pm$ 40.7} & \second{1.81 $\pm$ 0.05} \\
& & FAME & 128.0 $\pm$ 24.4 & 607.8 $\pm$ 107.6 & 1.67 $\pm$ 0.05 \\
\rowcolor{gray!15} \cellcolor{white} & \cellcolor{white} & & \textbf{53.8 $\pm$ 2.9} & \textbf{226.9 $\pm$ 9.0} & \\
\rowcolor{gray!15} \cellcolor{white} & \cellcolor{white}\multirow{-4}{*}{3} & \multirow{-2}{*}{\textbf{Ours}} & {\scriptsize \textbf{(↓53.7\%)}} & {\scriptsize \textbf{(↓33.8\%)}} & \multirow{-2}{*}{\textbf{1.81 $\pm$ 0.04}} \\
\cmidrule{2-6}

& & RACER & \second{89.3 $\pm$ 27.5} & \second{372.2 $\pm$ 65.5} & \second{1.80 $\pm$ 0.03} \\
& & FAME & 102.1 $\pm$ 13.3 & 640.5 $\pm$ 65.3 & 1.74 $\pm$ 0.05 \\
\rowcolor{gray!15} \cellcolor{white} & \cellcolor{white} & & \textbf{43.0 $\pm$ 2.0} & \textbf{243.9 $\pm$ 8.2} & \\
\rowcolor{gray!15} \cellcolor{white}\multirow{-12}{*}{\shortstack{Open-plan\\Office}} & \cellcolor{white}\multirow{-4}{*}{4} & \multirow{-2}{*}{\textbf{Ours}} & {\scriptsize \textbf{(↓51.8\%)}} & {\scriptsize \textbf{(↓34.5\%)}} & \multirow{-2}{*}{\textbf{1.82 $\pm$ 0.06}} \\
\midrule

& & RACER & \second{140.7 $\pm$ 20.3} & \second{420.2 $\pm$ 29.3} & \second{1.64 $\pm$ 0.07} \\
& & FAME & 163.7 $\pm$ 15.8 & 478.7 $\pm$ 45.1 & 1.63 $\pm$ 0.02 \\
\rowcolor{gray!15} \cellcolor{white} & \cellcolor{white} & & \textbf{89.5 $\pm$ 2.7} & \textbf{282.7 $\pm$ 21.4} & \\
\rowcolor{gray!15} \cellcolor{white} & \cellcolor{white}\multirow{-4}{*}{2} & \multirow{-2}{*}{\textbf{Ours}} & {\scriptsize \textbf{(↓36.4\%)}} & {\scriptsize \textbf{(↓32.7\%)}} & \multirow{-2}{*}{\textbf{1.70 $\pm$ 0.04}} \\
\cmidrule{2-6}

& & RACER & 114.5 $\pm$ 21.5 & \second{473.7 $\pm$ 52.4} & \second{1.71 $\pm$ 0.05} \\
& & FAME & \second{105.8 $\pm$ 22.6} & 508.2 $\pm$ 80.0 & 1.65 $\pm$ 0.02 \\
\rowcolor{gray!15} \cellcolor{white} & \cellcolor{white} & & \textbf{72.5 $\pm$ 3.8} & \textbf{304.6 $\pm$ 44.6} & \\
\rowcolor{gray!15} \cellcolor{white} & \cellcolor{white}\multirow{-4}{*}{3} & \multirow{-2}{*}{\textbf{Ours}} & {\scriptsize \textbf{(↓31.5\%)}} & {\scriptsize \textbf{(↓35.7\%)}} & \multirow{-2}{*}{\textbf{1.72 $\pm$ 0.02}} \\
\cmidrule{2-6}

& & RACER & \second{83.7 $\pm$ 14.3} & \second{495.0 $\pm$ 53.6} & \second{1.69 $\pm$ 0.03} \\
& & FAME & 96.3 $\pm$ 13.3 & 586.9 $\pm$ 63.0 & 1.67 $\pm$ 0.05 \\
\rowcolor{gray!15} \cellcolor{white} & \cellcolor{white} & & \textbf{52.8 $\pm$ 7.1} & \textbf{324.2 $\pm$ 32.9} & \\
\rowcolor{gray!15} \cellcolor{white}\multirow{-12}{*}{\shortstack{Octa\\Maze}} & \cellcolor{white}\multirow{-4}{*}{4} & \multirow{-2}{*}{\textbf{Ours}} & {\scriptsize \textbf{(↓36.9\%)}} & {\scriptsize \textbf{(↓34.5\%)}} & \multirow{-2}{*}{\textbf{1.70 $\pm$ 0.03}} \\

\bottomrule
\end{tabular}%
}

\vspace{0.2cm}
{\raggedright\footnotesize\textit{Note: All results are reported as mean $\pm$ std. Best and second-best results are shown in \textbf{black bold} and \second{gray bold}, respectively. Percentages in parentheses denote ours' improvement over the second-best method.}\par}
\vspace{-0.30cm}
\end{table}

\subsection{CP-guided Planning for Single UAV}
\label{sec:cp_planning}

Upon receiving its task list, each UAV executes CP-guided hierarchical planning to efficiently cover the assigned regions. Inspired by \cite{zhang2025falcon}, we formulate the global CP planning over the allocated task units as an Asymmetric Traveling Salesman Problem (ATSP). To ensure motion feasibility and smoothness, we incorporate velocity consistency into the edge cost rather than relying solely on path length. For a searched path $\mathcal{P}_{ij}=\{\mathbf{p}_0, \mathbf{p}_1, \cdots, \mathbf{p}_K\}$ between target positions $\mathbf{p}_0$ and $\mathbf{p}_K$, the segment-wise traversal time $T_{ij}$ with a velocity-consistency penalty is defined as:
\begin{equation}
T_{ij}=\frac{l_{ij}}{v_m} + 
\sum_{k=0}^{K-1}\Bigg(\frac{(v_m-|\hat{v}_k|)^2}{2v_ma_m}+\frac{2|\hat{v}_k|}{a_m}H(-\hat{v}_k)
\Bigg),
\end{equation}
where $l_{ij}$ is the path length derived from the connectivity graph, $v_m$ and $a_m$ are the maximum velocity and acceleration, and $H(\cdot)$ is the Heaviside step function. The term $\hat{v}_k$ represents the scalar projection of the initial velocity $\mathbf{v}_k$ along the direction $\mathbf{p}_k \rightarrow \mathbf{p}_{k+1}$:
\begin{equation}
\hat{v}_k=\mathbf{v}_k \cdot \frac{\mathbf{p}_{k+1}-\mathbf{p}_k}{\|\mathbf{p}_{k+1}-\mathbf{p}_k\|},
\end{equation}
where the initial velocity \(\mathbf{v}_k\) is defined as \(\mathbf{v}_0=\mathbf{v}_1\), 
\begin{equation}
\mathbf{v}_k = v_m \cdot \frac{\mathbf{p}_{k}-\mathbf{p}_{k-1}}{\|\mathbf{p}_{k}-\mathbf{p}_{k-1}\|}, \, k \geq 1.
\end{equation}

For local planning, we adopt the method in \cite{zhou2023racer}. We incrementally update frontiers, and cluster them using Principal Component Analysis. For each frontier cluster, we uniformly sample candidate viewpoints around the cluster centroid, and select the viewpoint covering the most frontier voxels as the viewpoint representative. Under the guidance of the global CP, we compute the local exploration tour within the target task unit via a variant of TSP with fixed start and end points, where the start point is the UAV’s current position and the end point is the vertex of the next task unit along the CP. Finally, minimum-time B-spline trajectory generation converts this discrete sequence into a smooth, dynamically feasible flight trajectory, incorporating explicit inter-UAV collision-avoidance constraints for safe coordination.

\section{EXPERIMENTS}

\subsection{Simulation Experiments}
\subsubsection{\textbf{Implementation Details}}
We set \( r_{\text{comm}} = 5\ \text{m} \), \(\sigma_Q = 1.1\) in Eq. (\ref{eq:capacity}) and \( \lambda_c = 1.2 \) in Eq. (\ref{eq:contiguity penalty}). The UAV dynamic motion limits are specified as: \( v_{\text{max}} = 2.0\ \text{m/s} \), \( \omega_{\text{max}} = 2.0\ \text{rad/s} \), and \( a_{\text{max}} = 2.0\ \text{m/s}^2 \). All simulations are conducted in MARSIM \cite{kong2022marsim}, where each UAV is equipped with a Livox MID360 LiDAR, with a perception range of 10 m for volumetric mapping. The CVRP and TSP are solved using the Lin-Kernighan-Helsgaun Solver \cite{helsgaun2017extension}. All the aforementioned modules run on an Intel Core i5-12400F CPU.

\subsubsection{\textbf{Benchmark Comparison}}
We evaluate \OURS\ in 3 scenarios: \textit{Cubicle Office} ($30 \times 30 \times 5 \text{ m}^3$), \textit{Open-plan Office} ($35 \times 30 \times 5 \text{ m}^3$), and \textit{Octa Maze} ($35 \times 35 \times 5 \text{ m}^3$). We compare our approach against SOTA decentralized multi-UAV exploration methods, RACER \cite{zhou2023racer} and FAME \cite{bartolomei2023fast}. Exploration efficiency is measured by exploration time, total path length, and average flight velocity. Results averaged over 10 trials per scenario are presented in Fig.~\ref{fig:benchmark} and Table~\ref{tab:benchmark}.

Across all scenarios, \OURS\ consistently outperforms both baselines, reducing the exploration time by 43.8\%, 50.7\% and 34.9\% and the total path length by 31.9\%, 33.7\% and 34.3\% in the \textit{Cubicle Office}, \textit{Open-plan Office} and \textit{Octa Maze}, respectively. Furthermore, it maintains a comparable flight velocity while achieving the lowest standard deviations in exploration time and path length, demonstrating superior robustness and stability.
In particular, the significant improvements in the \textit{Cubicle Office} and \textit{Open-plan Office} stem from the inherent limitations of both baselines. FAME relies on greedy, instantaneous local-frontier decisions, causing severe redundant exploration during communication failures. Meanwhile, RACER's topology-agnostic grid decomposition improperly merges disconnected regions (causing detours in the \textit{Open-plan Office}) and retains unreachable enclosed areas as valid tasks (yielding unreasonable allocations in the \textit{Cubicle Office}). In contrast, our connectivity-aware representation isolates disconnected components and filters out unreachable regions. In addition, our contiguity-driven allocation penalizes non-adjacent assignments via graph-based constraints, encouraging spatially contiguous tasks.

As the UAV team scales up, \OURS\ demonstrates robust scalability by effectively balancing exploration-time reduction and path-length growth. When scaling from 2 to 4 UAVs, our method achieves an average time reduction of 35.8\%. Although this relative scaling effect appears comparable to the baselines (RACER 32.9\%, FAME 36.8\%), it is primarily because \OURS\ already achieves exceptionally short times with just 2 UAVs. The true scalability advantage of our method emerges from suppressing redundant traversal: \OURS\ confines the average path length growth to merely 15.9\%, whereas RACER and FAME suffer from 18.0\% and 22.6\% inflation, respectively. This indicates that \OURS\ successfully harnesses additional UAVs without the severe path redundancy exhibited by the baselines.

\begin{table}[!t]
\vspace{0.0cm}
\centering
\caption{Ablation on core modules.}
\vspace{-0.10cm}
\label{tab:ablation_modules}
\renewcommand{\arraystretch}{1.1}
\setlength{\tabcolsep}{6pt}
\begin{tabular}{lcc}
\toprule
\textbf{Method} &
\textbf{Exploration Time (s)} &
\textbf{Path Length (m)} \\
\midrule
Ours       & \textbf{43.0} & \textbf{243.9}     \\
No-Con     & 51.7          & 329.6              \\
No-Graph   & 90.1          & 391.4              \\
\bottomrule
\end{tabular}
\vspace{-0.30cm}
\end{table}

\subsection{Ablation Studies}
To validate the effectiveness of two core modules in \OURS\ and evaluate its performance under communication constraints, we conduct two groups of ablation experiments in the challenging \textit{Open-plan Office} scenario with 4 UAVs.

\subsubsection{\textbf{Ablation of Core Modules}}
The graph-adjacency contiguity penalty is built upon our connectivity-aware task representation; without the connectivity graph, the penalty degenerates into a coarse grid-adjacency heuristic. Hence, it is not meaningful to ablate the connectivity-aware representation alone while keeping the same penalty definition. Instead, we evaluate two comparable variants, with results summarized in Table \ref{tab:ablation_modules}.
For \textbf{No-Con}, we retain the connectivity-aware task representation and the connectivity graph, but remove the graph-adjacency contiguity penalty from the CVRP cost matrix, solving task allocation using traversal cost only. For \textbf{No-Graph}, we discard the entire connectivity graph module and revert to a grid-based task representation by using the average position of unknown voxels in each uniform grid as a task unit.

Compared with the \OURS, \textbf{No-Con} increases exploration time by 20.2\% and total path length by 35.1\%, indicating that the contiguity penalty is crucial for maintaining spatially contiguous task sequences and avoiding inefficient cross-region traversal. \textbf{No-Graph} yields an even larger performance drop, confirming that our connectivity-aware task representation enables a more reasonable topological partition of the exploration space and reduces unnecessary detours caused by merging disconnected unknown regions into a single task unit.


\begin{figure}[!t]
  \vspace{0.0cm}
  \centering
  \includegraphics[width=0.96\columnwidth]{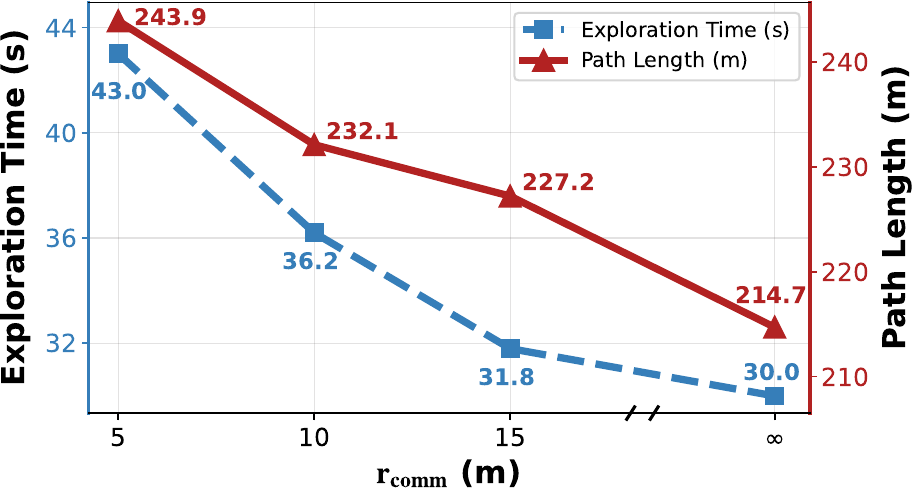}
  \vspace{-0.10cm}
  \caption{\textbf{Ablation on Communication Range}. Exploration time and total path length of \OURS\ under different communication ranges \(r_{\mathrm{comm}}\).}
  \label{fig:ablation_rcomm}
  \vspace{-0.30cm}
\end{figure}

\subsubsection{\textbf{Ablation of Communication Range}}
Although \OURS\ already achieves strong performance under the strict communication limit of \(r_{comm}=5\,\text{m}\), we further evaluate its adaptability under more relaxed communication conditions. Specifically, we test the system with \(r_{comm}\in\{5\,\text{m}, 10\,\text{m}, 15\,\text{m}\}\) and an idealized infinite-communication setting (\(r_{comm}=\infty\)). The results are shown in Fig.  \ref{fig:ablation_rcomm}.

As the communication range increases, exploration performance consistently improves and gradually approaches the upper bound given by infinite communication. Compared with the strict \(5\,\text{m}\) setting, the infinite-communication upper bound further reduces the exploration time by 30.2\% and shortens the total path length by 12.0\%. These results indicate that our framework not only remains stable and efficient under tight communication constraints, but also effectively leverages improved connectivity when available, demonstrating strong scalability and practical adaptability across varying communication conditions.

\subsection{Real-World Experiments}
To further demonstrate the robustness and effectiveness of \OURS, we conduct real-world experiments in an outdoor environment with 3 quadrotor UAVs. Each quadrotor UAV platform is equipped with an Intel NUC 13 for onboard computing, a Livox Mid-360 LiDAR for environment sensing, and an NxtPx4 for flight control. The state estimation module is based on FAST-LIO2 \cite{xu2022fast}, and the maximum flight velocity of all UAVs is set to \( 1.0\ \text{m/s}\).

As shown in Fig.~\ref{fig:intro}, we perform real-world flight validations in two outdoor scenarios: an unstructured woodland with randomly distributed trees ($40 \times 31 \times 3 \text{ m}^3$) and a structured teaching building with pillars ($48 \times 20 \times 3 \text{ m}^3$). The UAVs explored the scenarios in $69 \text{ s}$ and $52 \text{ s}$, respectively. The experimental results validate the feasibility of \OURS, demonstrating its capability to achieve efficient collaborative exploration through online task allocation. More details are provided in the submitted video.

\vspace{-0.0cm}
\section{CONCLUSIONS}

We present \OURS, a decentralized multi-UAV exploration framework that enhances task allocation flexibility and spatiotemporal contiguity under limited communication. By integrating a connectivity-aware task representation with a contiguity-driven allocation, \OURS\ achieves efficient exploration while reducing cross-region detours. Simulations and real-world experiments confirm that our method outperforms SOTA baselines, while demonstrating strong scalability regarding fleet size and communication range.

As future work, we will extend \OURS\ along two directions. First, we will augment passive communication with proactive connectivity maintenance, allowing the team to adaptively trade off exploration progress and coordination quality. Second, we will investigate tighter integration with localization-aware mapping and drift-resilient coordination to improve long-duration autonomy. These extensions are expected to further enhance robustness and scalability in dynamic, large-scale environments.

\section{ACKNOWLEDGMENT}
We would like to express our sincere gratitude to Yenan Wu, Yude Li, Junda Wu, Yichen Lai, Chenlin Gao, and Fangyu He from MASLab for their valuable assistance during the real-world experiments. We also sincerely thank Tengkai Zhuang from STARLab for his help in resolving the communication issues.

\vspace{-0.0cm}
\bibliographystyle{IEEEtran}
\bibliography{references}


\end{document}